\pdfoutput=1

\documentclass[11pt]{article}

\usepackage[]{acl}

\usepackage{times}
\usepackage{latexsym}

\usepackage{booktabs}
\usepackage{multirow}
\usepackage{enumitem}
\usepackage{graphicx}
\usepackage{amsmath}

\usepackage[T1]{fontenc}

\usepackage[utf8]{inputenc}

\usepackage{microtype}

\title{Probing Pre-Trained Language Models for Cross-Cultural \\Differences in Values}

\author{Arnav Arora \and Lucie-Aim\'{e}e Kaffee \and Isabelle Augenstein \\
  University of Copenhagen \\
  \texttt{\{aar, kaffee, augenstein\}@di.ku.dk}}

\date{}

\begin{document}
\maketitle

\begin{abstract}
Language embeds information about social, cultural, and political values people hold. Prior work has explored potentially harmful social biases encoded in Pre-trained Language Models (PLMs). However, there has been no systematic study investigating how values embedded in these models vary across cultures.
In this paper, we introduce probes to study which cross-cultural values are embedded in these models, and whether they align with existing theories and cross-cultural values surveys. We find that PLMs capture differences in values across cultures, but those only weakly align with established values surveys. We discuss implications of using mis-aligned models in cross-cultural settings, as well as ways of aligning PLMs with values surveys.
\end{abstract}

\section{Introduction}

A person's identity, values and stances are often reflected in the linguistic choices one makes~\citep{Jaffe2009StanceSP, norton-lang-identity}.
This is why, when language models are trained on large text corpora, they not only learn to understand language, but also pick up on a variety of societal and cultural biases~\citep{stanczak-etal-2021-gender-bias}. 
While biases picked up by the PLMs have a potential to cause harm when used in a downstream application, they may also serve as tools which provide insights into understanding cultural phenomena. Further, while studying ways of surfacing and mitigating potentially harmful biases is an active area of research, cultural biases and values picked up by PLMs remain understudied. Here, we investigate cultural values and differences among them picked up by PLMs through their pre-training on Web text. 

In a wide range of social science research fields, values are a crucial tool for understanding cross-cultural differences. As defined by \citet{rokeach2008understanding}, values are the ``core conceptions of the desirable within every individual and society'', i.e., the foundation for the beliefs guiding a persons actions and on a society level the base for the guiding principles. We would like to highlight the difference we make between \textit{values} and \textit{morals}. The former, as conceptualised in this work, is concerned with fundamental beliefs an individual or a group holds towards socio-cultural topics, whereas the latter entails making a judgement towards individual or collective right or wrong. For a discussion around the intersection of morality and PLMs, we point the reader to \citet{talat_word_2021}.
In this paper, we base our understanding of values across cultures on two studies: \citet{DBLP:conf/iwips/Hofstede05}, which defines 6 dimensions to describe cross-cultural differences in values, and the World Values Survey (WVS) \cite{wvs-7}.
Both surveys provide numerical value scores for several categories on a population level across different countries and regions and are widely used to understand cross-cultural differences in values.

PLMs are trained on large amounts of text from the Web and have shown to pick up on semantic, syntactical, factual and other forms of knowledge which allow them to perform well across several Natural Language Processing (NLP) tasks. Since multilingual PLMs are trained on text in many languages, they have the potential to pick up cultural values through word associations expressed in those languages which are embedded in the pre-training texts. We therefore measure whether cultural values embedded in multilingual PLMs are correlated with the ones provided by the surveys.
In Wikipedia, which is one of the primary sources of training data for multilingual PLMs, cross-cultural differences have been established \citep{DBLP:conf/icwsm/Miquel-RibeL19}, and analysed by \citet{DBLP:journals/jasis/HaraSH10} based on Hofstede's theory.

\begin{figure*}
    \centering
    \includegraphics[width=\textwidth]{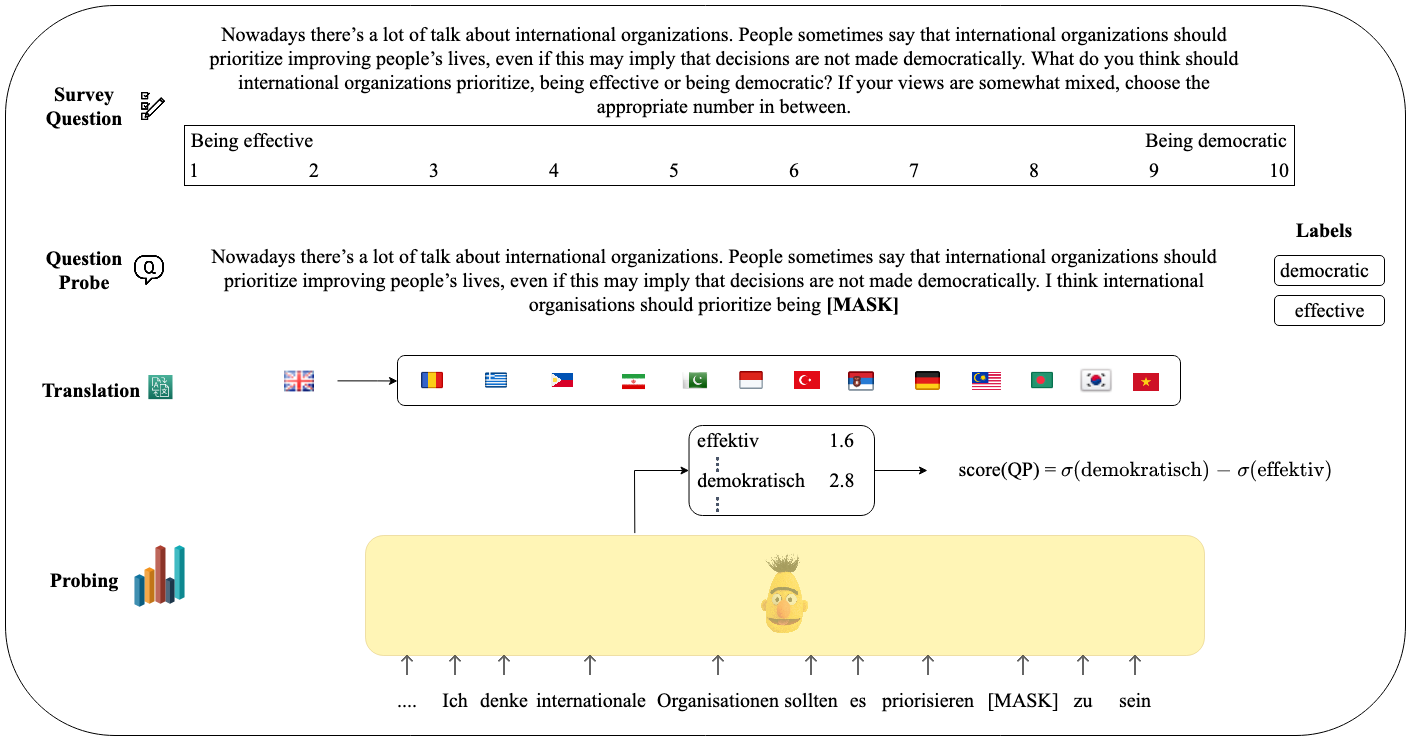}
    \caption{Figure outlining the experimental setting for the paper. We take the original survey questions (Section ~\ref{sec:val_surveys}), convert them into Question Probes and translate these into the target languages (Section ~\ref{para:multi_probes}) and run inference on the mask probes (Section ~\ref{subsec:experiments})}
    \label{fig:exp_setting}
\end{figure*}

In this paper, we explore the novel research question of whether PLMs capture cultural differences in terms of values across different language models. 
We probe PLMs using questions from the values surveys of both Hofstede's cultural dimensions theory and the World Values Survey.
We reformulate the survey questions to probes and extract the answers to evaluate whether language models can capture cultural differences based on their training data.
We focus on 13 languages, each of which is primarily geographically restricted to one country or region, to compare the results of the language models to the values surveys. The overall experimental setting for the paper is outlined in Figure~\ref{fig:exp_setting}.

Our work makes the following \textbf{contributions}\footnote{The Hofstede and World Values Survey probes used for our experiments can be found at \href{https://github.com/copenlu/value-probing}{https://github.com/copenlu/value-probing}.}: 
\begin{itemize}[noitemsep]
    \item We present the first study measuring cultural values embedded in large Pre-trained Language Models
    \item We propose a methodology for probing for values by converting survey questions to cloze style questions
    \item We conduct experiments across 13 languages with three multilingual language models (mBERT, XLM, and XLM-R), showing value alignment correlations with two large scale values surveys
    \item We present a discussion around potential implications of deploying these models in a multi-cultural context
\end{itemize}

\section{Related Work}

\paragraph{Expression and Norms}

Analysis of expression of identity and attitudes through language and its change has a long history in sociolinguistics~\cite{labov_social, trudgill2002}. More recently, studies have used NLP to computationally analyse this change on social media data~\citep{eisenstein-lex-change-2014, hovy-review-2015} and link it to external factors like socioeconomic status~\citep{abitbol-socioeco-2018} and demographics~\citep{jurgens-incoming-profiling}. This has also been done to analyse broader societal trends like temporal change in attitudes towards sexuality~\citep{ch-wang-jurgens-2021-using} and gender bias~\citep{sap-etal-2017-connotation,stanczak-gender-survey-2021}.

Further, there has been work on creating resources to analyse social norms and commonsense reasoning around them~\cite{forbes-etal-2020-social,emelin-etal-2021-moral, sap-etal-2020-social}.
\citet{hoover-2020-mft-corpus, roy-etal-2021-identifying} present work on extracting moral sentiment embedded in language using the Moral Foundation Theory. To diversify visually grounded reasoning across different cultures, ~\citet{liu-etal-2021-visually} introduce a multimodal multilingual dataset. 

While there has been work on investigating and embedding social and moral norms, understanding values and their variation in a cross-cultural context remains understudied in the literature. ~\citet{kiesel-etal-2022-identifying} provide a taxonomy of 54 values based on~\citet{schwartz_refining_2012} and provide a dataset and baselines for automatic value classification within the context of argument mining. The closest setup to ours would be one adopted by ~\citet{johnson-et-al-2022-ghost}. They qualitatively assess the text generated by GPT-3, an autoregressive language model, by prompting it with English texts with a clear embedded value. They find that the embedded values in the generated texts were altered to be more in line with dominant values of US citizens, possibly due to its training data. Our setup instead quantitatively measures whether cross-cultural differences in these values are preserved in multilingual language models when fed with the language spoken predominantly by people belonging to that culture.

\paragraph{Probing}
Probing has been extensively used as tool to study a variety of knowledge and biases picked up by PLMs. This can be syntactic~\citep{hewitt-manning-2019-structural}, semantic~\citep{vulic-etal-2020-probing}, numerical~\citep{wallace-etal-2019-nlp}, relational~\citep{petroni-etal-2019-language} or factual knowledge~\citep{jiang_x-factr_2020} picked up by PLMs. Probes can be created on both, at the word or sentence level~\citep{mosbach-etal-2020-interplay}.

Following work~\citep{caliskan-et-al-2017, garg-etal-2018-gender-stereotypes} on studying gender bias in word embeddings, a number of studies have built on it to similarly probe for social biases embedded in PLMs~\citep{may-etal-2019-measuring,guo-etal-2021-ceat,stanczak_quantifying_2021,ousidhoum-etal-2021-probing,de-vassimon-manela-etal-2021-stereotype, stanczak-etal-2021-gender-bias}. This can be done using cloze-style probing for measuring at an intra-sentence level~\citep{nadeem_stereoset_2021} or using pseduo-log likelihood~\citep{salazar-etal-2020-masked} based scoring~\citep{nangia-etal-2020-crows}. There are downsides to both approaches, the former potentially introduces unintended bias based on the tokens in the input probe while the latter assumes that all masked tokens are statistically independent~\citep{Kaneko_Bollegala_2022}. We choose the former since the probes in our case are carefully worded by social scientists with the explicit aim to extract bias towards a certain set of values.

To the best of our knowledge, there is no existing work on probing for values embedded in PLMs in a comparative cultural context. 

\section{Value Probing}
In this paper, we explore how PLMs capture differences in values across cultures, and whether those differences reflect the ones found in values across cultures at large.
To compare the PLMs' encodings of values, we compare them with established surveys capturing cross-cultual differences in values, namely Hofstede's cultural dimensions theory and the World Values Survey (WVS) (Section~\ref{sec:val_surveys}).
We transform the survey questions introduced in those surveys for compatibility with PLMs by reformulating them semi-automatically to convert them into probes (Section~\ref{sec:probe_gen}).
Then we translate these created probes from English into 13 geographically localised languages to conduct cultural value probing across 13 cultures (Section~\ref{para:multi_probes}).
Finally, we assess the variance in cross-cultural values embedded in PLMs and compare the probing results to the established values surveys in Section~\ref{sec:results}.

We investigate the following \textbf{research questions} as a first step to exploring this novel area of probing cross-cultural differences in values:

\begin{itemize}[noitemsep]
    \item[\textbf{RQ1}] Do PLMs capture diversity across cultures for the established values? 
    \item[\textbf{RQ2}] Are there similarities in the embedded values across different PLMs?
    \item[\textbf{RQ3}] Do values embedded in PLMs align with existing values surveys? 
\end{itemize}

\section{Values Surveys}
\label{sec:val_surveys}

We base our work on previous studies on how values differ across cultures. 

As these are central to a number of research fields including political science, psychology, sociology, behavioral economics, cross-cultural research among others, there are a large number of studies utilising the scores provided by these values surveys. Among the most common ones are Hofstede's cultural dimensions theory and the World Values Survey. These studies build on the body of work in different fields: Hofstede's theory is derived from management studies \citep{hofstede1984culture}, while the WVS was developed in the field of political science \citep{inglehart2006inglehart}. Both studies have since been widely used across fields. 

\subsection{Hofstede's Cultural Dimensions Theory}
\label{subsec:hof_survey}
Hofstede started his surveys of cross-cultural differences in values in 1980. This first survey \citep{hofstede1984culture} included 116,000 participants from 40 countries (extended to $111$ countries and regions in the 2015 version) working with IBM, and created 4 cultural dimensions, which were subsequently extended to 6 cultural dimensions that are also used in this paper. 
These 6 dimensions are: Power Distance (pdi), Individualism (idv), Uncertainity Avoidance (uai), Masculinity (mas), Long-term Orientation (lto), Indulgence (ivr). 
The full survey contains $24$ questions.
Each dimension is calculated using a formula defined by Hofstede using 4 of the questions in the survey, see Appendix~\ref{app:hofstedeformuals}.
Hofstede shows the influence that culture has on values by defining distinctly different numerical values in those 6 dimensions for the cultures observed.
While critics of Hofstede's cultural dimensions theory point out, among others, the simplicity of the approach of mapping cultures to countries and question the timeliness of the approach \citep{nasif1991methodological}, this model of representing values is now a foundation for a large body of work on cross-cultural differences in values \citep{jones2007hofstede}.

\subsection{World Values Survey (WVS)}
\label{subsec:wvs_survey}

The World Values Survey 
(WVS, \citet{wvs-7}) collects data on peoples' values across cultures in a more detailed way than Hofstede's cultural dimensions theory. The survey started in 1981 and is conducted by a nonprofit organisation, which includes a network of international researchers. It is conducted in \textit{waves}, to collect data on how values change over time. The latest wave, wave 7, ran from 2017 to 2020.
Compared to the European Values Study\footnote{\url{https://europeanvaluesstudy.eu/}}, WVS targets all countries and regions, and includes $57$ countries. 
While Hofstede's cultural dimensions theory aggregates the findings of their survey into the 6 cultural dimensions, WVS publishes the results of their survey per question. Those are organised in 13 categories: (1) Social Values, Attitudes and Stereotypes, (2) Happiness and Well-being, (3) Social Capital, Trust and Organisational Membership, (4) Economic Values, (5) Corruption, (6) Migration, (7) Security, (8) Postmaterialist Index, (9) Science and Technology, (10) Religious Values, (11) Ethical Values and Norms, (12) Political Interest and Political Participation, (13) Political Culture and Regimes.

We exclude categories (4) and (8) for the experiments in this study. This was done due to the nature of questions asked in these categories, for which it was not straightforward to design mask probes without loss of information.

\citet{inglehart2006inglehart}, who established WVS, further defines the \textit{Inglehart–Welzel cultural map}, which processes the surveys and defines two dimensions in relation to each other:  traditional versus secular-rational values and survival versus self-expression values, and summarise values for countries on a scatter plot describing these dimensions. In the following, we only use the previously mentioned 11 categories and leave an analysis based on the Inglehart–Welzel cultural map for future work.

\section{Probe Generation}
\label{sec:probe_gen}
In order to make the surveys compatible with language models, we reformulate the survey questions to cloze-style question probes \cite{taylor1953cloze,conf/nips/HermannKGEKSB15} that we can then perform masked language modelling inference on. Since this is the task PLMs were trained on, we argue it is a suitable methodology to measure embedded cultural biases in these models.
\paragraph{Hofstede's Cultural Dimensions}
Based on the English survey questions, the questions in the survey are manually reformulated to question probes (QPs). This is done analogously to iterative categorisation, in which a set of possible labels ($ y_{i}^+, y_{i}^-$) corresponding to either end of the response options available in the survey are defined, which are the words the language models are probed for. The sentences are then reformulated to probes, and the labels masked.
Those labels are based on the answers of the original survey, for instance, the original question like \textit{have sufficient time for personal or home life} with answer options consisting of different degrees of \textit{importance}, the probe is reformulated to \textit{Having sufficient time for personal or home life is [MASK].}, where \textit{[MASK]} should be replaced by \textit{important} or  \textit{unimportant}. 
\[ \text{QPs} = \{W_i, y_{i}^+, y_{i}^-\}\]

where $W_i$ is the masked probe and $y_{i}^+$ and $y_{i}^-$ are the set of labels.
There are a total of $24$ questions with repeating labels.

\paragraph{World Values Survey}
Analogous to the probes created from the Hofstede survey, we create probes from the English questionnaire of the WVS. As there are more questions than for Hofstede ($238$ in total), there are also a larger number of labels to replace and a higher variety of question types.

\paragraph{Multilingual Probes}
\label{para:multi_probes}

To probe across several languages, we follow a semi-automatic methodology for translating the created probes in English to the target language. We use a translation API\footnote{\url{https://cloud.google.com/translate}} that covers all target languages. We translate each QP from English into the target language with the [MASK] token replaced by the label words [$y_{i}^+, y_{i}^-$] in order to maintain grammatical structure and aid the translation API.
One challenge of cross-cultural research is information loss when translating survey questions \citep{nasif1991methodological,hofstede1984culture}. Therefore we opted for this approach rather than reformulating the translated survey questions by Hofstede. However, we would like to highlight the shortcomings of machine translation which have poor performance on low resource languages and has the potential to introduce additional biases. For the purpose of these experiments however, since the question probes are relatively simple sentences, we found the machine translations to be of high quality. We conducted an evaluation of our machine translated probes, the details for which can be found in the Appendix~\ref{sec:mt_eval}.
The target labels [$y_{i}^+, y_{i}^-$] for each QP are then translated individually as single words (e.g. \textit{important} is translated from English to the German \textit{wichtig}), followed by lowercased string matching to check if the translated label can be found and replaced in the translated probe.
If the target label cannot be found directly in the translated probe due to differences in word choice, we use a cross-lingual word aligner~\citep{DBLP:conf/eacl/DouN21} to align the English probe and its translated version. With this approach, we identify the label word to be replaced with the mask token. 
If both approaches yield no result, the token is manually replaced in the target sentence based on the authors' language understanding and using online translators.
\begin{table}[]
    \centering
    \begin{tabular}[]{llr}
        \toprule
        Country & Language & Wikipedia size\\
        \midrule
        Romania     &        Romanian (ro) & 428,330\\
        Greece      &        Greek (el) & 207,647\\
        Pakistan    &        Urdu (ur) & 168,587\\
        Iran        &        Farsi (fa) & 872,240\\
        Philippines &        Tagalog (tl) & 43,145\\
        Indonesia   &        Indonesian (id) & 618,395\\
        Germany     &        German (de) & 2,675,084\\
        Malaysia    &        Malay (ms) & 356,937\\
        Bangladesh  &        Bengali (bn) & 119,619\\
        Serbia      &        Serbian (sr) & 656,627\\
        Turkey      &        Turkish (tr) & 475,984\\
        Vietnam     &        Vietnamese (vi) & 1,270,712\\
        South Korea &        Korean (ko) & 582,977\\
        \bottomrule
    \end{tabular}
    \caption{Mapping of countries (cultures) to languages used throughout this paper, including number of articles per Wikipedia language as of March 2022.}
    \label{tab:languages_countries_mapping}
\end{table}
\paragraph{Language Selection}
\label{lang_selection}
In total, we investigate $13$ languages, mapped to one country each as outlined in Table~\ref{tab:languages_countries_mapping}, according to criteria further detailed below.
One of the limitations of this one-to-one mapping is that the languages are spoken in wider regions and not specifically in one country (disregarding also e.g. diaspora communities). 
This allows for the closest match to the values theories we work with, which operate on a country level.
The definition of culture by country has been criticised by, e.g., \citet{nasif1991methodological}.

We select the languages as follows: We first include the countries covered in both the surveys of WVS and Hofstede. We limit to languages which are official languages of the countries observed in the studies of both WVS and Hofstede. 
We further select languages for which the distribution of speakers is primarily localized to a country or relatively narrow geographical region.
To ensure the language models will be able to have (potentially) sufficient amount of training data, from the set of languages, only those are selected which have at least 10,000 articles on Wikipedia.

\section{Methodology}

\subsection{Models}
We conduct the probing experiments on three widely used multilingual PLMs: the multi-lingual, uncased version of BERT base (mBERT)~\citep{devlin2019bertpretrainingdeepbidirectional}, the 100 language, MLM version of XLM~\citep{conneau-xlm}, and the base version of XLM-RoBERTa (XLM-R)~\citep{conneau-etal-2020-unsupervised} available in the Transformers~\citep{wolf-etal-2020-transformers} library. mBERT was trained with a Masked Language Modelling (MLM) and Next Sentence Prediction objective, on Wikipedia articles in 102 languages with the highest number of articles on them. The XLM model builds on top of mBERT, only using the MLM objective but with modifications to the selection and truncation of training text fed to the model at each training step. It was also trained on Wikipedia texts, including 100 languages.
The XLM-R model uses the RoBERTa architecture~\citep{Liu2019RoBERTaAR} and is trained with an MLM objective on 2.5 TB of filtered CommonCrawl corpus data in 100 languages. It shows strong multilingual performance across a range of benchmarks and is commonly used for extracting multilingual sentence encodings. 

\subsection{Mask Probing}
\label{subsec:experiments}

For each model $M$, we run inference on the created cloze-style question probes (QPs, described in Section ~\ref{sec:probe_gen}) using an MLM head producing the log probabilities for the [MASK] tokens in the QPs over the entire vocabulary $V$ of the respective model: $logP_{M}(w_i,t|W_{i}^{\backslash t}, \Theta_M) \in {R}^{|V|}$, where $t$ is the position of the [MASK] token in the text $W_i \in QP$, and $\Theta_M$ are the parameters of the corresponding Language Model $M$. Since the survey respondents have to answer the questions with a choice between a range of values, for instance 1-10 with 1 representing \textit{democratic} and 10 representing \textit{effective}, in order to replicate a similar setting with PLMs, we 
subtract the predicted logit for the response label with the highest score $w_i^+$ with the predicted logit for the lowest score $w_i^-$. 
This normalises the predicted logits for the responses on opposing ends of the survey question and is then used as a score for that question. \[logP_{M}(w_i) = logP_{M}(w_i^+) - logP_{M}(w_i^-)\]

Finally, in order to collapse the World Values Survey responses per \textit{category}, within which many questions have different scales, we normalize the aggregate survey responses per the corresponding question scale, so that $y_{i, c} \in [0,1], c \in C$. We then take the mean of the responses across all the questions of the category to arrive at the aggregated score of the category for each country: $y_i = \frac{1}{|C|}\sum_{c \in C}{y_{i,c}}  \in [0, 1]$. 

\subsection{Evaluation}

We calculate Spearman's $\rho$ -- a rank correlation coefficient between the values predicted by the language models and values calculated through the surveys: $\rho(logP_{M}(w_i,t|W_{i}^{\backslash t}, \Theta_M), y_i) $. For the World Values Survey, we do this per question, as well as per category. For Hofstede, we limit this calculation to value level correlations due to lack of access to individual or aggregate survey response data per question.\footnote{We calculate the scores for the values based on the formula provided at \url{https://www.laits.utexas.edu/orkelm/kelmpub/VSM2013_Manual.pdf}, see Appendix~\ref{app:hofstedeformuals}.} We further calculate correlations per country. 
Spearman's $\rho$ works on relative predicted ranks to each variable, ignoring the individual predicted values. Our choice of using a rank correlation was motivated by the fact that we are working with population level aggregate responses and our aim of assessing whether language models pick up on relative differences in values across cultures, rather than on exact values.

\section{Results}\label{sec:results}
\subsection{RQ1: Model Predictions} 
\begin{figure}[!t]
    \centering
    \includegraphics[width=\columnwidth]{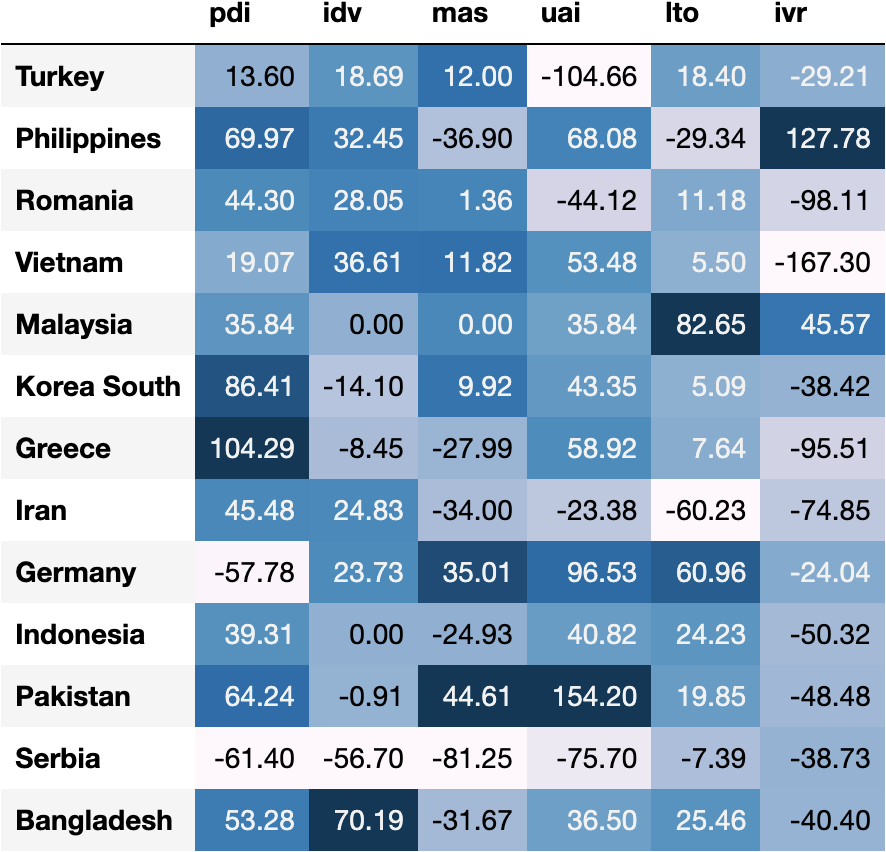}
    \caption{Heatmap of scores predicted per value for XLM-R mask probing on Hofstede's survey questions}
    \label{fig:hof_vals_xlmr}
\end{figure}

We show the predicted scores for the XLM-R model in Figure~\ref{fig:hof_vals_xlmr}
. As is clear from the figure, there are substantial differences in the predicted scores for the cultural dimensions across cultures. On average, scores for \textit{power distance} (pdi) are high, whereas ones for \textit{masculinity} (mas) and \textit{indulgence} (ivr) are relatively low. The predicted logits suggest bias towards \textit{Greece} and \textit{South Korea} as places with high power distance, \textit{Pakistan}, \textit{Germany} as more masculine. \textit{Indulgence} (ivr) has the lowest scores across all values with only \textit{Phillippines} and \textit{Malaysia} having positive values, indicating high restraint in these cultures according to the model predictions.

\begin{figure*}[ht!]
    \centering
    \includegraphics[width=\textwidth]{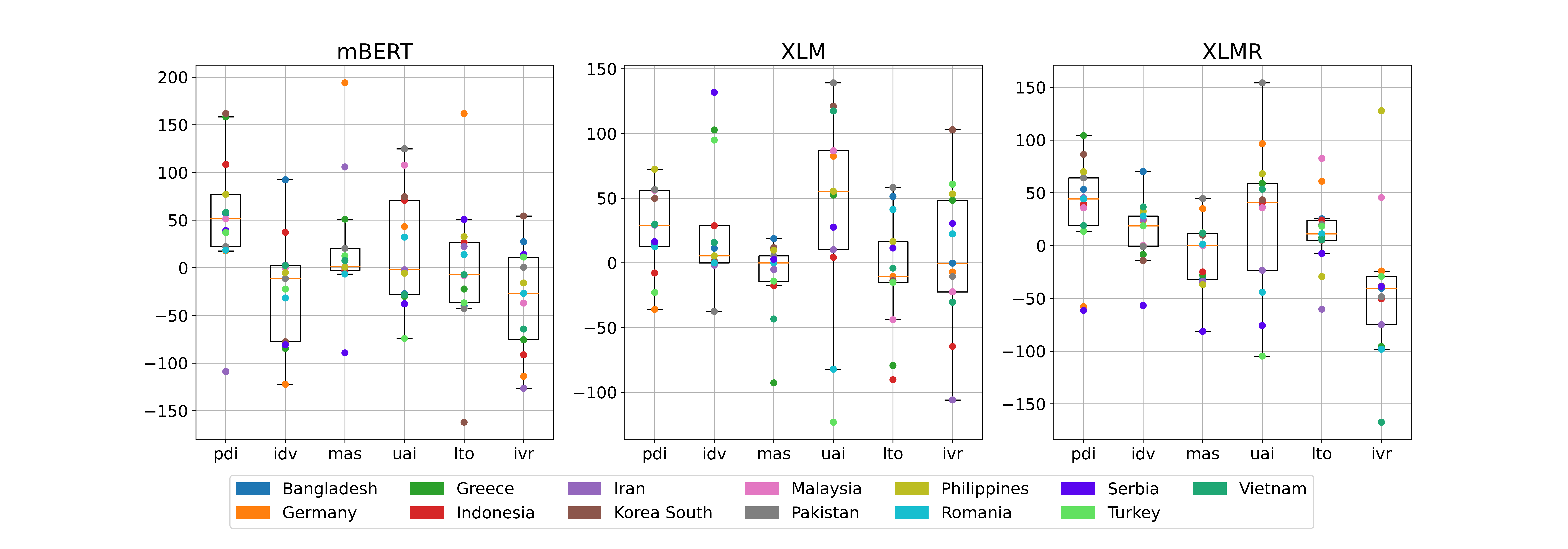}
    \caption{Scatter plots with quartiles of predicted value scores on Hofstede's survey questions for each of the three models.}
    \label{fig:var_hof}
\end{figure*}

\begin{figure*}[ht!]
    \centering
    \includegraphics[width=\textwidth]{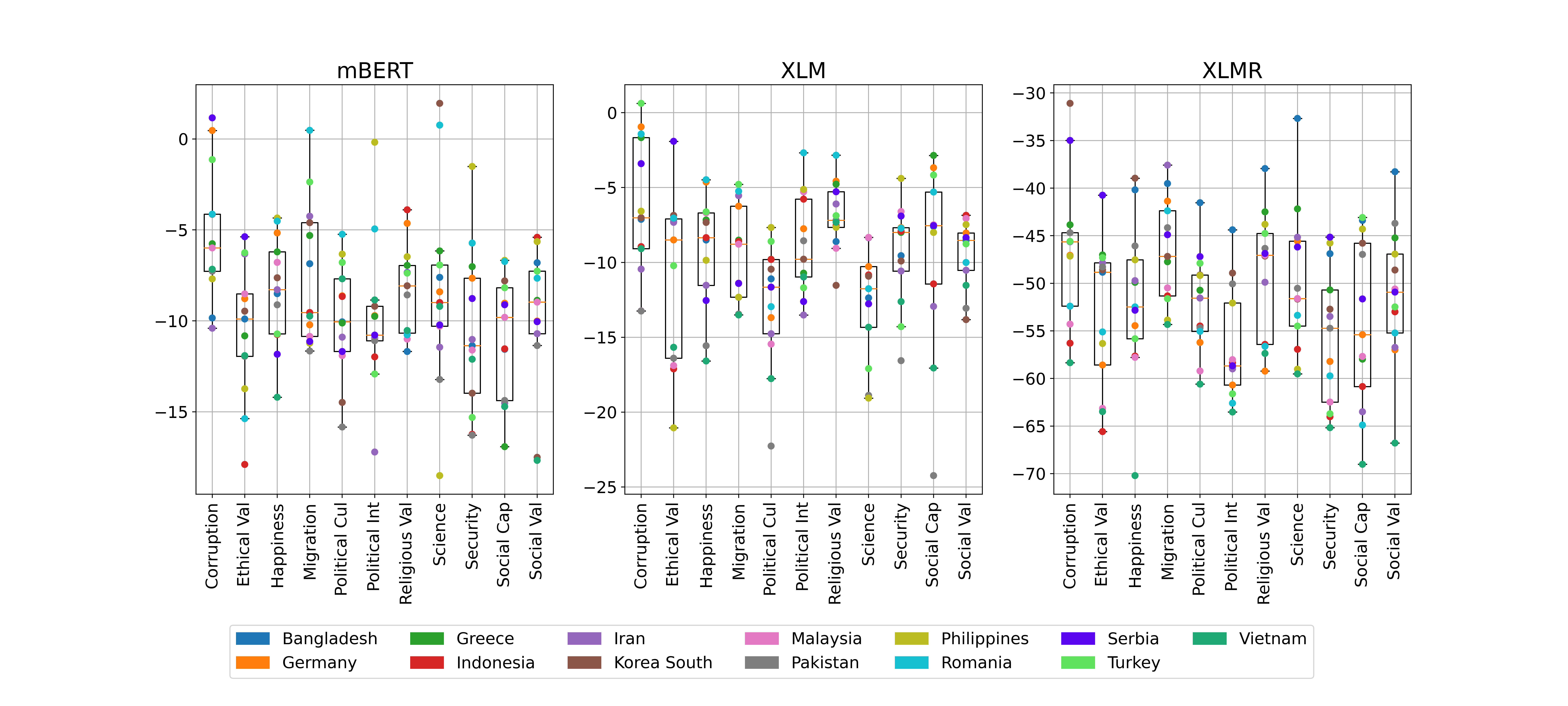}
    \caption{Scatter plots with quartiles of predicted value scores on WVS questions  for each of the three models.}
    \label{fig:var_wvs}
\end{figure*}

To understand whether LMs can preserve cross-cultural differences in values, we plot the results of the probing for Hofstede's and WVS' survey in Figures~\ref{fig:var_hof} and~\ref{fig:var_wvs} respectively. As is visible in these plots, there is a variety in the values, i.e., the models seem to place different importance on different values across cultures, displaying cross-cultural differences in the values.
We quantify these differences among the prediction scores by testing for statistical significance between the model's predictions by culture, seeing how they capture cross-cultural differences. For XLM-R's predictions for the WVS,  42.31\% of the country pairs have a statistically significant difference, meaning the model preserves cross-cultural differences. For the other two models, the share of significantly different country pairs are 51.28\% and 46.15\% for mBERT and XLM respectively.
For XLM-R's predictions of Hofstede's survey, only 10.26\% of cultures have $p<=0.05$. 
For the other two models, the share of significantly different country pairs are none and 6.41\% for mBERT and XLM respectively.
We attribute these low percentages to the fact that we conduct the test over the six value dimensions only, while it is on over 200 questions for WVS. 

\subsection{RQ2: Model Agreement}

To further study whether scores across values and categories are consistent across the three models, we check for correlation between the predicted scores between the three models and outline them in Tables~\ref{tab:mod_agreement_hof} and~\ref{tab:mod_agreement_wvs}. We can see that predictions are inconsistent across the models, indicating differences in the embedded cross-cultural values. mBERT and XLM share the same architecture and are both trained on Wikipedia, yet the correlations across values are low, indicating the large effect that relatively minor changes to the model training can have on the cultural values picked up by the model.

\begin{table}[t]
    \centering
    \begin{tabular}[]{lccc}
        \toprule
        {} &  {XLM/} &  {XLM/} &  {mBERT/} \\
        {} &  {mBERT} &  {XLM-R} &  {XLM-R} \\
        \midrule
        pdi &  0.44 & 0.68* & 0.48\\
        idv & -0.14 & -0.22 & 0.55\\
        mas & -0.41 & -0.14 & 0.43\\
        uai &  0.49 & 0.65* & 0.42\\
        lto & -0.05 & -0.12 & -0.15\\
        ivr &  0.67* & 0.39 & 0.3\\
        \bottomrule
    \end{tabular}
    \caption{Pairwise correlations in predictions for mask probing on Hofstede's values questions. Statistically significant values with $ p \leq 0.05 $ are marked with *}
    \label{tab:mod_agreement_hof}
\end{table}

\begin{table}[t]
    \centering
    \begin{tabular}[width=\columnwidth]{lccc}
        \toprule
        {} &  {XLM/} &  {XLM/} &  {mBERT/} \\
        {} &  {mBERT} &  {XLM-R} &  {XLM-R} \\
        \midrule
        Corruption   &   0.57* & 0.53 & 0.44\\
        Ethical Va  &  0.61* & 0.79* & 0.47\\
        Happiness &  0.49 & 0.24 & 0.63*\\
        Migration       &  0.16 & 0.44 & 0.25\\
        Political Cu &  0.38 & 0.65* & 0.57*\\
        Political In &  0.6* & 0.81* & 0.75*\\
        Religious & 0.09 & -0.31 & 0.05\\
        Science &  0.51 & 0.24 & 0.21\\
        Security        &  0.49 & 0.77* & 0.83*\\
        Social Cap & 0.21 & 0.4 & 0.42\\
        Social Val  &  0.61* & 0.27 & 0.68*\\
        \bottomrule
    \end{tabular}
    \caption{Pairwise correlations in model predictions for mask probing on WVS questions. Statistically significant values with  $p \leq 0.05$ are marked with *}
    \label{tab:mod_agreement_wvs}
\end{table}

\subsection{RQ3: Alignment with Surveys}

Finally, we investigate whether the models' predictions for the values questionnaire are consistent with existing values survey scores.

\paragraph{Hofstede}
We outline the results of correlations between each of the models' predictions for mask probing per value in Table~\ref{tab:model_hof_corr_mask}. We find no statistically significant alignment between the models' predictions and survey value scores provided by Hofstede, but given the low sample size, this is to be expected~\citep{sullivan2012using}. We find weak correlations among some of the values between the models' predicted scores and the values survey suggesting the disparity in cultural values outlined by Hofstede and the ones picked up by PLMs.

\begin{table}[h]
    \centering
    \begin{tabular}[]{l|c|c|c}
        \toprule
        {} &  {mBERT} &  {XLM} &  {XLM-R} \\
        \midrule
        ivr &       -0.44 &      0.07 &       0.38  \\
        idv &       -0.38 &     -0.04 &       0.21 \\
        mas &        0.37 &      0.09 &      -0.07 \\
        uai &       -0.30 &     -0.30 &      -0.22 \\
        pdi &        0.25 &      0.16 &      -0.11 \\
        lto &        0.02 &     -0.01 &       0.23 \\
        \bottomrule
    \end{tabular}
    \caption{Correlation per value between mask prediction scores and Hofstede's values survey. Statistically significant values with p <= 0.05 are marked with *}
    \label{tab:model_hof_corr_mask}
\end{table}


\paragraph{WVS}

Table~\ref{tab:model_wvs_corr} similarly shows the correlations between the models' predicted scores and the World Values Survey scores per category. Here too, we find no statistically significant correlation between the predicted and the survey scores outlining the difference in values picked up by the language models and those quantified in the surveys.

We also check for per country correlations between the predicted scores and data from both values surveys, these are shown in Tables~\ref{tab:model_hof_corr_country} and~\ref{tab:model_wvs_corr_country} in the Appendix.

\begin{table}[h]
    \centering
    \begin{tabular}{l|c|c|c}
        \toprule
        {} &  {mBERT} &  {XLM} &  {XLM-R} \\
        \midrule
        Science                            &        0.50 &      0.09 &       0.19 \\
        Security                                           &        0.38 &     -0.22 &       0.09 \\
        Social Val           &       -0.34 &     -0.30 &      -0.07 \\
        Political Cul                      &        0.29 &      0.15 &      -0.05 \\
        Political Int     &        0.25 &      0.02 &       0.10 \\
        Migration                                          &        0.19 &      0.26 &       0.21 \\
        Social Cap &        0.17 &      0.06 &      -0.38 \\
        Religious Val                                   &        0.14 &      0.13 &      -0.37 \\
        Corruption                                         &        0.07 &      0.12 &       0.12 \\
        Happiness                            &       -0.07 &      0.37 &       0.07 \\
        Ethical Val                           &        0.04 &     -0.02 &       0.03 \\
        \bottomrule
    \end{tabular}
    \caption{Correlation per question between masked prediction scores and WVS. Statistically significant values with p <= 0.05 are marked with *}
    \label{tab:model_wvs_corr}
\end{table}

\section{Discussion}

Our experiments show that there are sizable differences in the cultural values picked up by the different multilingual models which are widely used for a number of language tasks, even when they are trained on data from the same source. This is in line with previous results~\citep{stanczak-etal-2021-gender-bias} and hints at the sensitivity of model design, training choices, and their downstream effect on model biases. While the values picked up by the models vary across cultures, the bias in the models is not in line with values outlined in existing large scale values surveys. This is an unexpected result since PLMs are known to pick up on biases present in language data that they are trained on~\cite{rogers-etal-2020-primer, stanczak-gender-survey-2021}. Further, values are known to be expressed in language~\citep{norton-lang-identity}. Hence, language models should pick up on and reflect cultural differences in values expressed in different languages based on their training text. A lack of such reflection points to possible shortcomings in representation learning when it comes to multilingual language models. There could be a number of reasons for this. One possible reason is the lack of diversity in multilingual training data. Wikipedia articles in different languages are written by a small subset of editors that are not representative of the populations in those countries. Further, large scale corpora like CommonCrawl over-represent the voices of people with access to the Internet, which in turn over-represents the values of people from those regions~\citep{bender-etal-2021-parrots}. Such a bias being present in GPT-3 was explored by~\citet{johnson-et-al-2022-ghost} who show that LMs trained on Web text end up reflecting the biases of majority populations. Other work also shows that pre-training text contains substantial amounts of toxic and undesirable content even after filtering~\citep{luccioni-viviano-2021-whats}. This highlights the need for including more diverse and carefully curated sources of data which are culturally sensitive and representative, in order for the models to better reflect the cultural values of those populations.
In this pursuit, knowledge gathered from the field of Cultural Anthropology in following an \textit{emic} methodology can prove to be useful~\citep{hansen-2020-emic}. 
\citet{DBLP:conf/emnlp/JosephSGGMAL21} suggest that people express themselves differently online on Twitter compared to survey responses. This is another potential reason for this mis-alignment. 

PLMs are used for a variety of different NLP tasks in different countries and hence to accommodate the usage of people from diverse backgrounds and cultures, it is not just important to have linguistic and typological diversity in training data, but also cultural diversity~\citep{hershcovich2022challenges}. Having such a form of cultural knowledge is desirable for a number of real-world tasks including QA systems, dialogue systems, information retrieval. Further, a lack of such faithful representation could lead to unintended consequences during the deployment of such models such as models imposing a set form of normative ethics over a diverse population that may not subscribe to it~\citep{talat_word_2021,johnson-et-al-2022-ghost}. This could also lead to models not being culturally sensitive and embedding harmful stereotypes~\citep{nadeem_stereoset_2021}. 
Recently, work has been done on trying to align models with human values~\citep{hendrycks2021aligning,solaiman-et-al-2021-palms}. While this may seem like a good idea at a first glance, also in light of the arguments presented above, some cultural values are harmful to portions of society, e.g. high levels of masculinity, which is connected to misogynistic language and perpetuating gender biases. Thus, when working with cultural values, an auditing system~\citep{raji-et-al-2020-auditing} with these value systems in mind and one that takes into account the downstream use case should be employed.

\section{Conclusion}
In this study, we propose a methodology for probing of cultural values embedded in multilingual Pre-trained Language Models and assessing differences among them. We measure alignment of these values amongst the models and with existing values surveys. 
We find that PLMs capture marked differences in values between cultures, though these in turn are only weakly correlated with values surveys. Alongside training data, we discuss the impact training and modelling choices can have on cultural bias picked up by the models.
We further discuss the importance of this alignment when developing models in a cross-cultural context and offer suggestions for more inclusive ways of diversifying training data to incorporate these values. 

\section{Ethical Considerations}
The ethical considerations for our work mostly relate to the limitations; there are a variety of unintended implications of equating a language and a country, such as misrepresentation of communities, and disregarding minority and diaspora communities. However, we believe it is the closest approximation possible when comparing the surveys used in this work and LMs. 
Further, the surveys have been criticised; particularly Hofstede's cultural dimensions theory has been deemed too simplistic \cite{jackson2020legacy}. This could lead also to simplistic assumptions when probing an LM. 

We address these problems by including the WVS, another widely used survey, in our study. 
Due to these limitations, we believe that further studies and applications of our approach should be done with these limitations in mind. Particularly the simplification of cultural representation by both our approach as well as the original surveys might impact communities negatively. Such misrepresentation can have a disproportionate impact and exacerbate the marginalisation of minority communities or subcultures.

\section{Acknowledgements}
This research was co-funded by a DFF Sapere Aude research leader grant under grant agreement No 0171-00034B.

\section{Limitations}
There are several limitations of our approach in trying to assess cultural diversity and alignment of the values picked up by PLMs. While our methodology of probing models using Cloze style questions gives us some insight into token level biases picked up by the language models, it is limited in its approach to only show static and extrinsic biases at inference time using output probabilities. There are intrinsic measures for quantifying bias, but those do not always correlate with extrinsic measures~\cite{goldfarb-tarrant-etal-2021-intrinsic}. In order to make the experimental setting more robust and clearly demonstrate signs of embedded cultural bias, we performed experiments with an extended set of synonyms for each label word. However, this turned out to be non-trivial for a number of reasons. First, replacing synonyms in place of original words rarely results in grammatical sentences. Second, it is not always possible to find multiple synonyms of words in the same sense as the label words across the languages used in our study. Third, even when synonyms do exist, they are often multi-word expressions, which makes them incompatible with our experimental setting where a single word needs to be masked. \\
As discussed earlier, a major limitation that comes with quantifying cultural values is the mapping of countries to cultures and in our case, also to languages. Since this is an imperfect mapping, it is a difficult task to accurately quantify and assess cultural bias and values embedded in the models. We partially addressed this by restricting our study to languages which are mostly geographically restricted to one country. This is a limitation faced by cross-cultural research in general, where countries are often used as surrogates for cultures \citep{nasif1991methodological}. 

Finally, surveys and aggregate responses are also imperfect tools to evaluate and quantify cultural disparity, though the best ones currently in use. They are tasked with collapsing individual values into a set of questions. Individuals answering those questions from different backgrounds may perceive the questions differently. Further, there are several confounding factors affecting the survey responses and problems relating to seeing populations as a monolithic homogeneous whole.
While these limitations pose important questions around how one should be careful in interpreting these values, we believe our study makes important contributions and provides a first step in assessing alignment between PLMs and cultural values, which we argue is necessary for models to faithfully work in a cross-cultural context.

\bibliography{tacl2021,anthology}
\clearpage

\appendix

\section{Translation quality}
\label{sec:mt_eval}

To assess the quality of translated probes, we conduct human evaluations of a sample of the output of the machine translator. We randomly select 3 probe questions from the Hofstede values survey and 23 probe questions from the World Values Survey representing 10\% of the total probes. We then provide the original probe questions in English as well as their translations to annotators and assess the following two characteristics of the translations:
\begin{itemize}[noitemsep]
    \item \textit{Grammaticality}: describes the correctness of the sentence standing alone, independent of the English sentence, in terms of obeying grammatical rules
    \item \textit{Meaning}: describes how adequate the translation is for further reuse. We specifically want to know here, how correct the sentence is in relation to the English sentence. This could be also understood as the overall quality of the translation.
\end{itemize}
For each of the 26 probe questions, we ask the annotators to rate the sentence on the above listed characteristics across a 1-5 Likert scale. All annotators had at least a university level education, working proficiency of English, and were native speakers of the corresponding languages. We perform this annotation for 6 out of the 13 languages due to resource constraints. We provide the averaged scores for both the characteristics for each language in Table~\ref{tab:mteval}.
The annotators on average across languages rate the meaning characteristic of the machine translated probes to be 4.73. This indicates the high degree to which the translations preserve the meaning of the sentences from the English probes. The grammaticality of the probes on average was rated to be 4.64. While lower than the value for the preserved meaning of the English sentence, the sentences were found to have very good grammar as well. The very high scores across the meaning characteristic of the translations suggest that for most of the probes, the translations were of high quality. 
\begin{table}[]
    \centering
    \begin{tabular}[]{lcc}
        \toprule
        Language & Grammaticality & Meaning\\
        \midrule
        German & 4.85 & 4.88\\
        Indonesian & 4.62 & 4.77\\
        Urdu & 4.54 & 4.88\\
        Serbian & 4.88 & 5\\
        Greek & 4.58 & 4.80\\
        Bengali & 4.38 & 4.04\\
        \hline
        Average & 4.64 & 4.73\\
        
        \bottomrule
    \end{tabular}
    \caption{Averaged human evaluation scores on a 1-5 Likert scale for grammaticality and preserved meaning of the machine translated probes for a sample of languages used in this study}
    \label{tab:mteval}
\end{table}

\section{Models and Compute} 
 All models were run in Python using PyTorch~\citep{NEURIPS2019_9015} and the Transformers library~\citep{wolf-etal-2020-transformers}. When speaking about XLM-R, mBERT, XLM we refer to the models with the names xlm-roberta-base,  bert-base-multilingual-uncased, xlm-mlm-100-1280 respectively.
Since only inference was performed for probing the models, the experiments were run on a single NVIDIA Titan RTX GPU for less than 1 hour.

\section{Ablations}
\label{sec:ablation}
\subsection{Label logit subtraction}

To eliminate the possibility of lack of correlation due to subtraction of logit for label token with the lower response score in the survey question from the one with higher response score (Section~\ref{subsec:experiments}), we calculate correlations with just the high response label token $y_{i}^+$. We report our results for Hofstede in Table~\ref{tab:hof_ablation_lab} and WVS in Table~\ref{tab:wvs_ablation_lab}. Similarly, we calculate value correlations for just the low response label and report them in Table~\ref{tab:hof_ablation_ant} and Table~\ref{tab:wvs_ablation_ant} for Hofstede and WVS respectively.

\begin{table}[h!]
    \centering
    \begin{tabular}{l|c|c|c}
        \toprule
        {} &  {mBERT} &  {XLM} &  {XLM-R} \\
        \midrule
        mas &        0.46 &    -0.11 &     -0.05  \\
        uai &        0.46 &     0.13 &      0.06  \\
        ivr &       -0.39 &     0.50 &      0.41  \\
        idv &       -0.38 &     0.51 &      0.12  \\
        pdi &        0.16 &    -0.00 &     -0.16  \\
        lto &       -0.13 &    -0.05 &     -0.02  \\
        \bottomrule
    \end{tabular}
    \caption{Correlation per dimension between mask prediction scores for the high response score label $y^+$ and Hofstede’s values survey. Statistically
significant values with p <= 0.05 are marked with *}
    \label{tab:hof_ablation_lab}
\end{table}

\begin{table}[h!]
    \centering
    \begin{tabular}{l|c|c|c}
        \toprule
        {} & {mBERT} &  {XLM} &  {XLM-R} \\
        \midrule
        Science  &        0.51 &      0.40 &       -0.13 \\
        Social Val           &       -0.44 &     -0.50 &       0.16  \\
        Political Cul                      &        0.43 &       0.33 &       0.08 \\
        Corruption        &        0.39 &     0.42 &       -0.11 \\
        Ethical                           &       -0.24 &      0.10 &         0.29 \\
        Religious            &       -0.16 &       -0.06 &         0.36  \\
        Migration                                          &        0.14 &         0.17 &        0.08\\
        Political Int     &        0.06 &      0.16 &       -0.21  \\
        Security       &       -0.06 &        -0.09 &       -0.12 \\
        Happiness                           &       -0.06 &      -0.09 &         0.21  \\
        Social Cap &       -0.06 &       -0.55* &       0.22 \\
        \bottomrule
    \end{tabular}
    \caption{Correlation per category between mask prediction scores for the high response score label $y^+$ and the WVS. Statistically
significant values with p <= 0.05 are marked with *}
    \label{tab:wvs_ablation_lab}
\end{table}

\begin{table}[]
    \centering
    \begin{tabular}{l|c|c|c}
        \toprule
        {} & {mBERT} &  {XLM} &  {XLM-R} \\
        \midrule
        Ethical            &        0.63* &      0.06 &      0.32 \\
        Security           &       -0.34 &     -0.05 &      -0.20 \\
        Religious          &       -0.27 &     -0.26 &       0.37 \\
        Social Val         &       -0.25 &     -0.61* &      0.15 \\
        Political Int      &       -0.13 &      0.28 &      -0.18 \\
        Migration          &        0.08 &      0.09 &      -0.19 \\
        Political Cul      &        0.06 &      0.12 &       0.08 \\
        Happiness          &       -0.03 &     -0.47 &       0.21 \\
        Corruption         &       -0.03 &      0.34 &      -0.20 \\
        Social Cap         &        0.01 &     -0.47 &       0.26 \\
        Science            &       -0.00 &     -0.40 &      -0.28 \\
        \bottomrule
    \end{tabular}
    \caption{Correlation per category between mask prediction scores for the low response score label $y^-$ and the WVS. Statistically
    significant values with p <= 0.05 are marked with *}
    \label{tab:wvs_ablation_ant}
\end{table}

\begin{table}[h]
    \centering
    \begin{tabular}{l|c|c|c}
        \toprule
        {} &  {mBERT} &  {XLM} &  {XLM-R} \\
        \midrule
        mas &         0.73* &         0.55* &     0.02  \\
        uai &         0.55* &        -0.39  &    -0.21  \\
        idv &         0.18  &         0.45  &    -0.08  \\
        ivr &        -0.16  &        -0.27  &    -0.11  \\
        lto &        -0.08  &        -0.06  &    -0.38  \\
        pdi &        -0.01  &         0.62* &     0.39  \\
        \bottomrule
    \end{tabular}
    \caption{Correlation per dimension between mask prediction scores for the low response score label $y^-$ and Hofstede’s values survey. Statistically significant values with p <= 0.05 are marked with *}
    \label{tab:hof_ablation_ant}
\end{table}

\begin{table}[t]
    \centering
    \begin{tabular}{l|c|c|c}
        \toprule
        {} &  {mBERT} &  {XLM} &  {XLM-R} \\
        \midrule
        Romania     &        0.93*&     -0.26  &       0.38 \\
        Pakistan    &        0.70 &      0.84* &       0.99* \\
        Greece      &        0.54 &     -0.09  &       0.49 \\
        Indonesia   &        0.54 &     -0.31  &       0.66 \\
        Vietnam     &        0.49 &     -0.14  &      -0.43 \\
        Serbia      &        0.37 &     -0.43  &      -0.31 \\
        Germany     &        0.26 &      0.23  &       0.60 \\
        Philippines &        0.26 &      0.54  &       0.20 \\
        Bangladesh  &       -0.20 &      0.58  &       0.23 \\
        Iran        &       -0.14 &      0.83* &       0.66 \\
        Turkey      &       -0.14 &     -0.83* &      -0.71 \\
        Malaysia    &       -0.09 &     -0.06  &       0.41 \\
        Korea South &        0.03 &     -0.03  &       0.54 \\
        \bottomrule
    \end{tabular}
    \caption{Correlation per country between mask prediction scores and Hofstede's values survey. Statistically significant values with p <= 0.05 are marked with *}
    \label{tab:model_hof_corr_country}
\end{table}

\begin{table}[h!]
    \centering
    \begin{tabular}{l|c|c|c}
        \toprule
        {} & {mBERT} & {XLM} & {XLM-R} \\
        \midrule
        Greece      &        0.78* &     -0.26 &       0.01 \\
        Philippines &        0.67* &      0.53 &       0.36 \\
        Turkey      &       -0.56  &     -0.34 &      -0.86* \\
        Malaysia    &        0.44  &      0.31 &       0.23 \\
        Bangladesh  &        0.43  &     -0.36 &       0.10 \\
        Vietnam     &        0.28  &      0.46 &       0.26 \\
        Iran        &       -0.24  &     -0.43 &      -0.09 \\
        Korea South &       -0.20  &     -0.36 &      -0.06 \\
        Romania     &        0.20  &      0.14 &      -0.18 \\
        Indonesia   &        0.18  &      0.34 &       0.03 \\
        Germany     &        0.13  &      0.09 &       0.06 \\
        Serbia      &        0.03  &     -0.01 &      -0.14 \\
        Pakistan    &       -0.01  &      0.17 &       0.23 \\
        \bottomrule
    \end{tabular}
    \caption{Correlation per country between masked prediction scores and World Values Survey. Statistically significant values with p <= 0.05 are marked with *}
    \label{tab:model_wvs_corr_country}
\end{table}

\section{Example probes}
\label{app:eg_probes}
In Table~\ref{tab:eg_probes}, we provide a sample of the question probes in English that are then translated to the different languages outlined in Section~\ref{sec:probe_gen}. 

\begin{table}[h!]
\small
 \begin{tabular}{p{2.5cm}p{8cm}p{1.5cm}p{1.5cm}}
\toprule
Value  & Probe & $y^+$ & $y^-$ \\
\midrule
Power Distance &  I [MASK] that one can be a good manager without having a precise answer to every question that a subordinate may raise about his or her work &  agree &  disagree \\
Individualism &  Having pleasant people to work with is [MASK] &  important &  unimportant \\
Masculinity &  Having a job respected by your family and friends is [MASK] &  important &  unimportant \\
Uncertainty Avoidance &  I feel [MASK] to be a citizen of my country &  proud &  ashamed \\
Long-term Orientation &  In my experience, subordinates are [MASK] afraid to contradict their boss (or students their teacher) &  never &  always \\
Indulgence &  All in all, I would describe the state of my health these days as [MASK] &  good &  bad \\
Corruption &  There is [MASK] corruption in my country &  abundant &  no \\
Economic Vals &  I [MASK] that competition is good &  agree &  disagree \\
Ethical Vals &  Government monitoring all emails and any other information exchanged on the internet should be [MASK] &  legal &  illegal \\
Happiness &  In the last 12 months, I or my family have [MASK] without cash income &  often &  never \\
Migration &  I [MASK] that the government should let anyone from other countries who wants to &  agree &  disagree \\
Political Cul &  Having the army rule is very [MASK] &  good &  bad \\
Political Int &  Attending peaceful demonstrations is something I have [MASK] done &  always &  never \\
Science &  I completely [MASK] that because of science and technology, there will be more opportunities for the next generation &  agree &  disagree \\
Security &  Drug sale in the streets is [MASK] in my neighbourhood &  frequent &  infrequent \\
Social Capital &  I am an [MASK] member of women's group &  active &  inactive \\
Social Vals &  It is [MASK] for me to have people who speak a different language as neighbours & undesirable &  desirable \\
\bottomrule
\end{tabular}

\caption{Examples of question probes in English reformulated from the original survey questions.}
    \label{tab:eg_probes}
\end{table}

\section{Hofstede Value Calculation}
\label{app:hofstedeformuals}
We calculate the value results for the probes based on \citet{hofstede1984culture} by using the formulas used in the original survey.\footnote{The formulas are provided along with the survey results and other information at \url{https://www.laits.utexas.edu/orkelm/kelmpub/VSM2013_Manual.pdf}.} The numbers following $m$ represent the index of the survey questions, $m$ stands for mean representing the mean survey question response for the answer to that question, $C$ is a constant that does not influence the comparison between countries.

\textbf{Power Distance} defined as "the extent to which
the less powerful members of organizations and institutions accept and expect that power is distributed unequally".
\[ pdi = 35(m07 - m02) + 25(m20 - m23) + C(pd)\]
\textbf{Individualism} measures "the degree to which people
in a society are integrated into groups".
\[
    idv = 35(m04 - m01) + 35(m09 - m06) + C(ic) 
\]
\textbf{Uncertainity Avoidance} measures "the extent to which a culture programs its members to feel either uncomfortable
or comfortable in unstructured situations".
\[
    mas = 35(m05 - m03) + 35(m08 - m10) + C(mf) 
\]
\textbf{Masculinity} index indicates "the nature of clearly distinct social and emotional gender roles in a society."
\[
    uai = 40(m18 - m15) + 25(m21 - m24) + C(ua) 
\]
\textbf{Long term orientation} Cultures with short-term orientation value "reciprocating social obligations, respect for tradition, protecting one’s ’face’, and personal steadiness and stability more".
\[
    lto = 40(m13 - m14) + 25(m19 - m22) + C(ls)
\]
\textbf{Indulgence} indicates "a society that allows relatively free gratification of basic and natural human desires related to enjoying life and having fun."
\[
    ivr = 35(m12 - m11) + 40(m17 - m16) + C(ir) 
\]

\end{document}